\documentclass[runningheads]{llncs}
\usepackage{graphicx}

\usepackage{amsthm}
\usepackage{amsmath}
\usepackage{mathtools}
\usepackage{booktabs}
\usepackage{xcolor}
\usepackage{colortbl}
\usepackage{multirow}
\usepackage[ruled]{algorithm2e}
\usepackage{amssymb}
\usepackage{pifont}
\usepackage{adjustbox}
\newcommand{\cmark}{\ding{51}}%
\newcommand{\xmark}{\ding{55}}%
\newcommand{\argmax}{\operatornamewithlimits{argmax}}
\usepackage{wrapfig}
\usepackage[misc,geometry]{ifsym}

\usepackage[colorlinks=True,
            linkcolor=red,
            anchorcolor=blue,  
            pagebackref,
            urlcolor=black,
            citecolor=blue,
            ]{hyperref}
\begin{document}
\title{Unsupervised Prototype Adapter \\for Vision-Language Models}
%
%\titlerunning{Abbreviated paper title}
% If the paper title is too long for the running head, you can set
% an abbreviated paper title here
%
\author{Yi Zhang\inst{1} \and
Ce Zhang\inst{1} \and
Xueting Hu\inst{1} \and
Zhihai He\inst{1, 2}\textsuperscript{(\Letter)}
}
\authorrunning{Y. Zhang et al.}
% First names are abbreviated in the running head.
% If there are more than two authors, 'et al.' is used.
%
\institute{Southern University of Science and Technology, Shenzhen, China \and
Pengcheng Laboratory, Shenzhen, China \\
\email{\{zhangyi2021,zhangc2019,huxt2022\}@mail.sustech.edu.cn}\\
\email{hezh@sustech.edu.cn}}

\maketitle              % typeset the header of the contribution
\begin{abstract}
Recently, large-scale pre-trained vision-language models (\textit{e.g.} CLIP and ALIGN) have demonstrated remarkable effectiveness in acquiring transferable visual representations.
To leverage the valuable knowledge encoded within these models for downstream tasks, several fine-tuning approaches, including prompt tuning methods and adapter-based methods, have been developed to adapt vision-language models effectively with supervision. 
However, these methods rely on the availability of annotated samples, which can be labor-intensive and time-consuming to acquire, thus limiting scalability.
To address this issue, in this work, we design an unsupervised fine-tuning approach for vision-language models called Unsupervised Prototype Adapter (UP-Adapter). Specifically, for the unannotated target datasets, we leverage the text-image aligning capability of CLIP to automatically select the most confident samples for each class. Utilizing these selected samples, we generate class prototypes, which serve as the initialization for the learnable prototype model. After fine-tuning, the prototype model prediction is combined with the original CLIP's prediction by a residual connection to perform downstream recognition tasks.
Our extensive experimental results on image recognition and domain generalization show that the proposed unsupervised method outperforms 8-shot CoOp, 8-shot Tip-Adapter, and also the state-of-the-art UPL method by large margins.

\end{abstract}

\begin{figure}[htbp!]
\includegraphics[width=\linewidth]{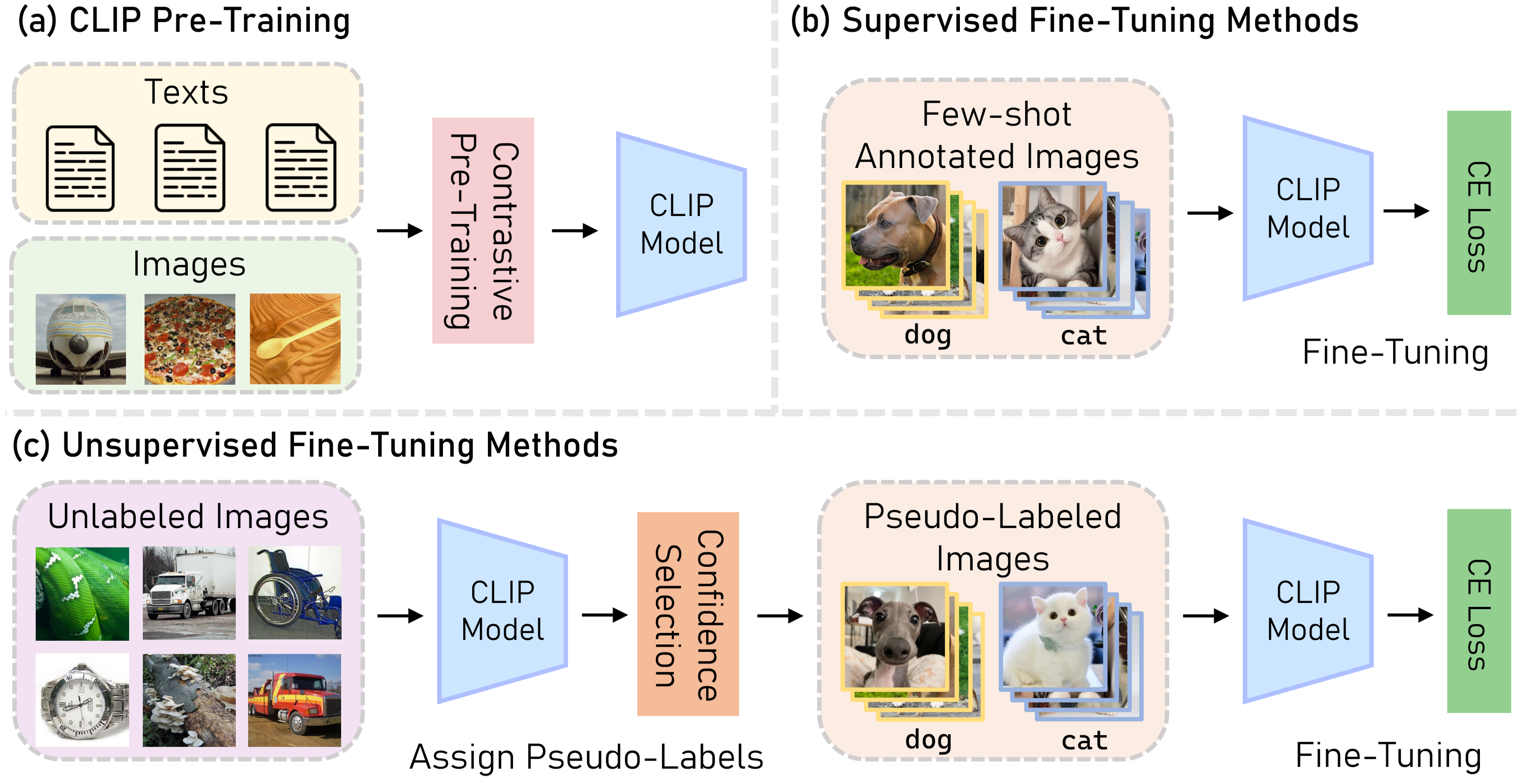}
\vspace{-23pt}
\caption{(a) CLIP pre-training process. (b) Typical supervised fine-tuning methods, \textit{e.g.}, CoOp \cite{zhou2022learning} and Tip-Adapter \cite{zhang2022tip}. (c) Our proposed unsupervised fine-tuning method.}
\label{fig:unsupervised}
\vspace{-11pt}
\end{figure}

\section{Introduction}
\label{sec:intro}
Transferring knowledge to unseen scenarios is a capability natural to humans yet challenging
for deep learning models to reproduce \cite{wang2020generalizing,zhou2022domain}. During the past decade, researchers have focused on designing sophisticated architectures and advanced algorithms to train visual recognition systems to predict accurately for a limited set of visual categories. However, this approach often fails to perform consistently on unseen classes unless the classifier is retrained on a more comprehensive dataset.

Recently, pioneering CLIP \cite{radford2021learning} has provided a new paradigm for generic visual recognition: pre-training, then fine-tuning. Under this paradigm, large-scale vision-language models are established based on pre-training on image-text pairs available on the Internet \cite{radford2021learning,yu2022coca,wang2022simvlm,huang2022idea}, then practitioners fine-tune these powerful models  to downstream tasks through supervised learning on annotated samples \cite{zhou2022learning,gao2021clip,zhang2022tip}. Illustrations of the CLIP pre-training process and fine-tuning process are shown in Fig. \ref{fig:unsupervised} (a) and (b). Those fine-tuned vision-language models have shown impressive performance on various vision tasks, such as image recognition \cite{radford2021learning,gao2021clip,zhou2022learning}, object detection \cite{shi2022proposalclip,du2022learning}, and image captioning \cite{li2021grounded,yao2021cpt}.

Although those supervised fine-tuning approaches, such as CoOp \cite{zhou2022learning} and Tip-Adapter \cite{zhang2022tip} have shown robust adaptation capabilities, these methods still require annotated samples, which are labor-intensive and time-consuming to obtain and limit the scalability. In this work, we provide an unsupervised fine-tuning approach as shown in Fig. \ref{fig:unsupervised} (c) to address this important issue. In our approach, we make use of the text-image aligning capability of CLIP to generate pseudo-labels for unlabeled samples. We then carefully select samples based on confidence scores for each class and generate class prototypes from these selections. These prototypes are utilized to initialize the prototype model, which can be significantly boosted by fine-tuning with a few training epochs. The prediction of the prototype model is integrated with the original CLIP's prediction to predict the final labels of images.  Our extensive experimentation in image recognition and domain generalization demonstrates significant performance improvements of our unsupervised method UP-Adapter compared to 8-shot CoOp, 8-shot Tip-Adapter, and the state-of-the-art unsupervised UPL method.

Our method has the following three \textit{unique novelties}:
(1) Leveraging the visual-textual correlation capability of CLIP, we address the challenge of requiring a substantial amount of annotated data.
(2) We have concurrently addressed the challenge of mitigating the substantial computational costs. For ImageNet, UPL needs 200 epochs and 15 hours to achieve an accuracy of 61.09\%, while our method achieves an accuracy of 63.58\% with only 20 epochs and 3 minutes.
(3) We are the first to explore the potential of the adapter-based fine-tuning method in the unsupervised setting. Our method works directly on the image feature, which can guarantee the preservation of prior knowledge in pre-trained CLIP.

\section{Related Work}
% In this section, we review  related works on vision-language pre-trained models, fine-tuning vision-language models, and self-training with pseudo-labeling.

\subsection{Large-Scale Pre-Trained Vision-Language Models}
Over the past few years, vision-language pre-trained models have emerged as a powerful approach to tackle various tasks that involve the interaction between visual and textual information \cite{radford2021learning}.  These models leverage large-scale datasets to learn representations that capture the semantic understanding of both images and their associated textual descriptions.
For example, CLIP \cite{radford2021learning} employs contrastive learning between the embeddings of 400 million carefully selected image-text pairs.
Researchers have demonstrated that large-scale pre-trained vision-language  models have great potential in visual representation learning and transfer learning, therefore can be utilized in several tasks such as image retrieval \cite{lu2019vilbert,duan2022multi}, visual question answering \cite{zhou2022unsupervised}, \textit{etc}.

\subsection{Adaptation Methods for Vision-Language Models}
To adapt vision-language models to various downstream tasks, researchers have developed a number of \textit{supervised} and \textit{unsupervised} approaches.

Recent \textit{supervised} methods typically try to adapt the models using few-shot samples  to inherit the well-learned prior knowledge of vision-language models to downstream tasks. These methods can be classified into two major directions: prompt tuning methods and adapter-based methods.
\textbf{Prompt tuning methods} focus on refining prompts and incorporating learnable context to extract task-relevant information from the encoded knowledge \cite{zhou2022learning,zhou2022conditional}. A notable example is CoOp \cite{zhou2022learning}, which optimizes the prompt context using learnable vectors. CoCoOp \cite{zhou2022conditional} builds upon CoOp by generating vectors conditioned on each image, addressing the challenge of generalizing to unseen classes. TPT \cite{shu2022tpt} enables the learning of adaptive prompts on the fly with just a single test sample. \textbf{Adapter-based methods} focus on tuning the text and image feature representations  directly \cite{zhang2022tip,zhang2023cross}. For example, CLIP-Adapter \cite{gao2021clip} designs a feature adapter to enhance traditional fine-tuning results. Building upon this, Tip-Adapter (Training-free CLIP-Adapter) \cite{zhang2022tip} achieves improved performance by establishing a key-value cache model from few-shot samples to perform fine-tuning. 

Another line of research delves into the \textit{unsupervised} setting where the labels for the target images are unavailable. For instance, UPL \cite{huang2022unsupervised} proposes to generate pseudo-labels for the unannotated images to convert this problem to a supervised problem. Some approaches also suppose the target images are inaccessible, and propose to adapt the vision-language models with only the class names (called ``name-only transfer"). For instance, VisDesc \cite{menon2023visual} queries large language models for concept descriptors for image classification, CHiLS \cite{novack2023chils} performs inference in the subclass space, while SuS-X \cite{udandarao2022sus} constructs a support set using text-to-image generation model or large-scale vision-language dataset.

\subsection{Self-Training with Pseudo-Labeling}
Semi-supervised learning has been extensively studied over several decades. Its primary objective is to develop a predictive model using a small set of annotated samples and a large set of unlabeled samples \cite{van2020survey}.
One notable approach within semi-supervised learning is self-training, which was initially proposed by Scudder \cite{scudder1965probability}. Self-training approaches typically leverage a model trained on annotated samples to generate pseudo-labels for unlabeled samples. 
% These pseudo-labeled training samples are then utilized together with the annotated samples to train a more robust model \cite{amini2022self}. 
% The self-training scheme has been widely adopted and has demonstrated impressive performance in various computer vision tasks, such as image recognition \cite{wu2020semi,xie2020self}, object detection \cite{li2020improving}, \textit{etc}. 
In the realm of fine-tuning vision-language models, the concept of self-training was initially introduced by UPL \cite{huang2022unsupervised}. UPL specifically utilizes the pre-trained CLIP model directly for pseudo-labeling and employs these pseudo-labeled samples to optimize the prompt representations. In this work, we aim to further explore the potential of self-training in fine-tuning vision-language models.

\section{Method}
\subsection{Background}

CLIP has two parallel encoders: an image encoder and a text encoder. For the image encoder, ResNet \cite{he2016deep} or Vision Transformer (ViT) \cite{dosovitskiy2020image} are commonly utilized to extract the visual features, while for the text encoder, we typically use a Transformer-based text. During the training phase, CLIP utilizes a contrastive loss function to facilitate the similarity between the features of the images and text in the embedding space. After training, the two modalities are aligned within a shared embedding space.

A CLIP model is denoted as $\{E_t, E_v\}$, where $E_t$ refers to the text encoder, and $E_v$ refers to the image encoder. In the image recognition task, a single test image $X_{test}$ belonging to a specific class $y$ is provided, where $X_{test}\in \mathbb{R} ^{C\times H\times W}$ and $y \in \mathbb{R} ^ K$ for a $K$-class classification problem. In zero-shot CLIP, each $y_i$ in the set $Y=\{y_1, y_2, \cdots, y_K\}$ is  concatenated with a pre-defined prompt such as $\rho =$ ``a photo of," to create textual inputs for different classes, denoted as $\{\rho; y_i\}$. Text features, $\{t_1, t_2, \cdots, t_K\}$, are generated by the text encoder $E_t$, where $t_i = E_t({\rho; y_i})$. Subsequently, we compute a cosine similarity score using each text feature $t_i$ is combined with the image feature $v=E_v(X_{test})$, which can be used to predict the probability of $X_{test}$ belonging to class $y_i$.
\begin{equation}
\label{eq-sim}
   \mathrm{sim}\left( t_i,v \right)=\frac{t_i \cdot v}{\Vert t_i \Vert  \Vert v \Vert}.
\end{equation}

The prediction probability on $X_{test}$ can be denoted by 
\begin{equation}
\label{eq-clip}
   p(y_i|X_{test})=\frac{\exp \left( \mathrm{sim}\left( t_i,v \right) /\tau \right)}{\sum\nolimits_{j=1}^K{\exp \left(\mathrm{sim}\left( t_j,v \right) /\tau \right)}}, 
\end{equation}
where $\tau$ refers to the temperature hyperparameter of the softmax function. The final predicted label $\hat{y}$ can be written by:

\begin{equation}
\hat{y} = \argmax_{y_i} p_{y_i}.
\label{eq:pseudo_label}
\end{equation}

\subsection{Unsupervised Prototype Adapter}

\begin{figure}[ht]
\vspace{-14pt}
\includegraphics[width=1.0\linewidth]{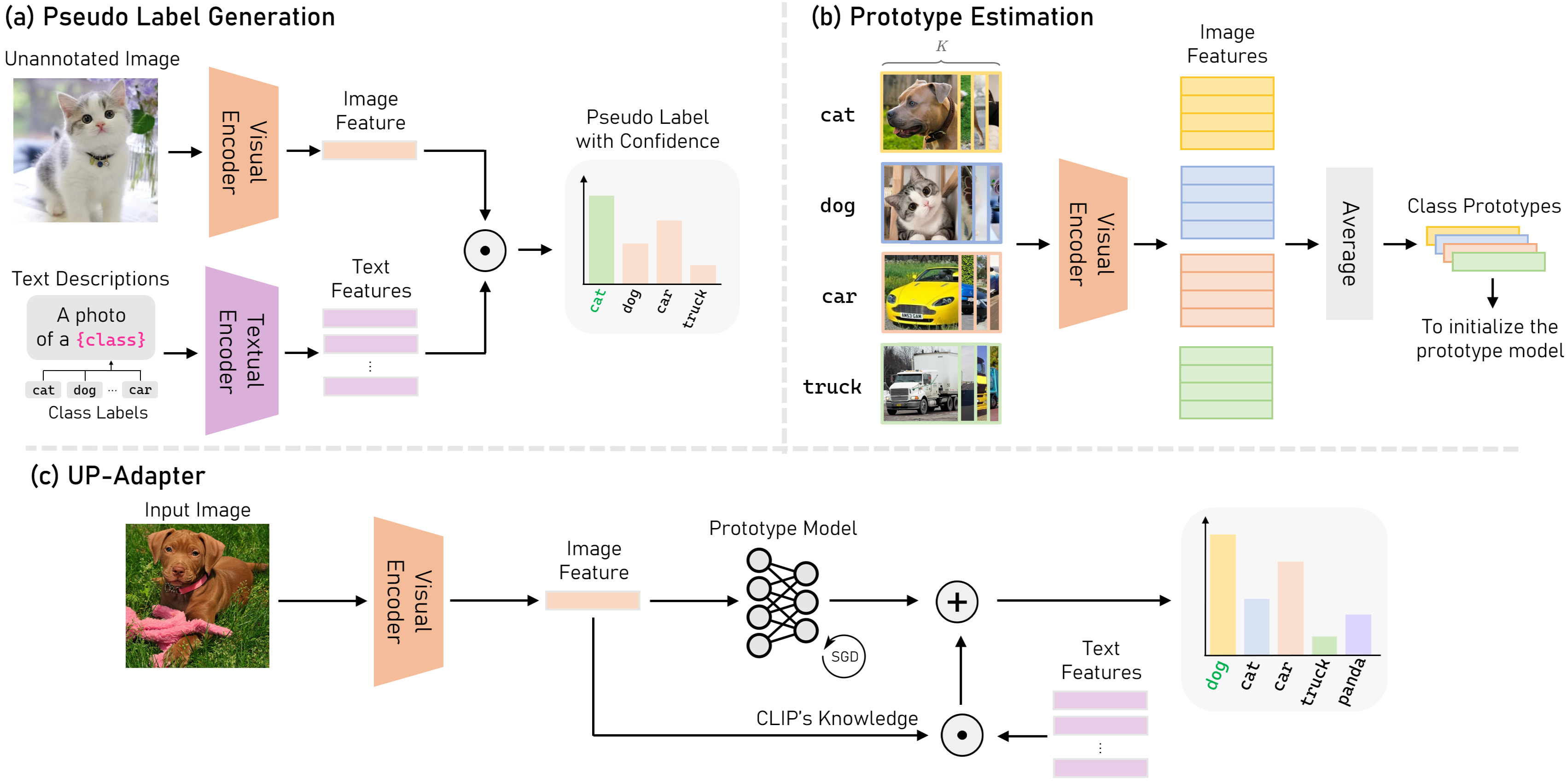}
\vspace{-20pt}
\caption{An overview of our proposed UP-Adapter method. (a) shows the pseudo-label generation process, for every class in the target dataset, we choose the most confident top-$K$ samples for subsequent optimization. (b) is the class prototype estimation process, for every class, we generate the features of 16 selected samples by image encoder, then average the features as the class prototype. (c) presents the overall process of our proposed UP-Adapter, we fuse the predictions of original pre-trained CLIP and the prototype model for final classification.}
\label{fig:overview}
\vspace{-15pt}
\end{figure}

In Fig. \ref{fig:overview}, we present an overview of our proposed  UP-Adapter method. The proposed UP-Adapter consists of 3 stages: pseudo-label generation, class prototype estimation, and prototype adapter.

\begin{figure}[htbp!]
% \vspace{-10pt}
\begin{center}
\includegraphics[width=0.83\linewidth]{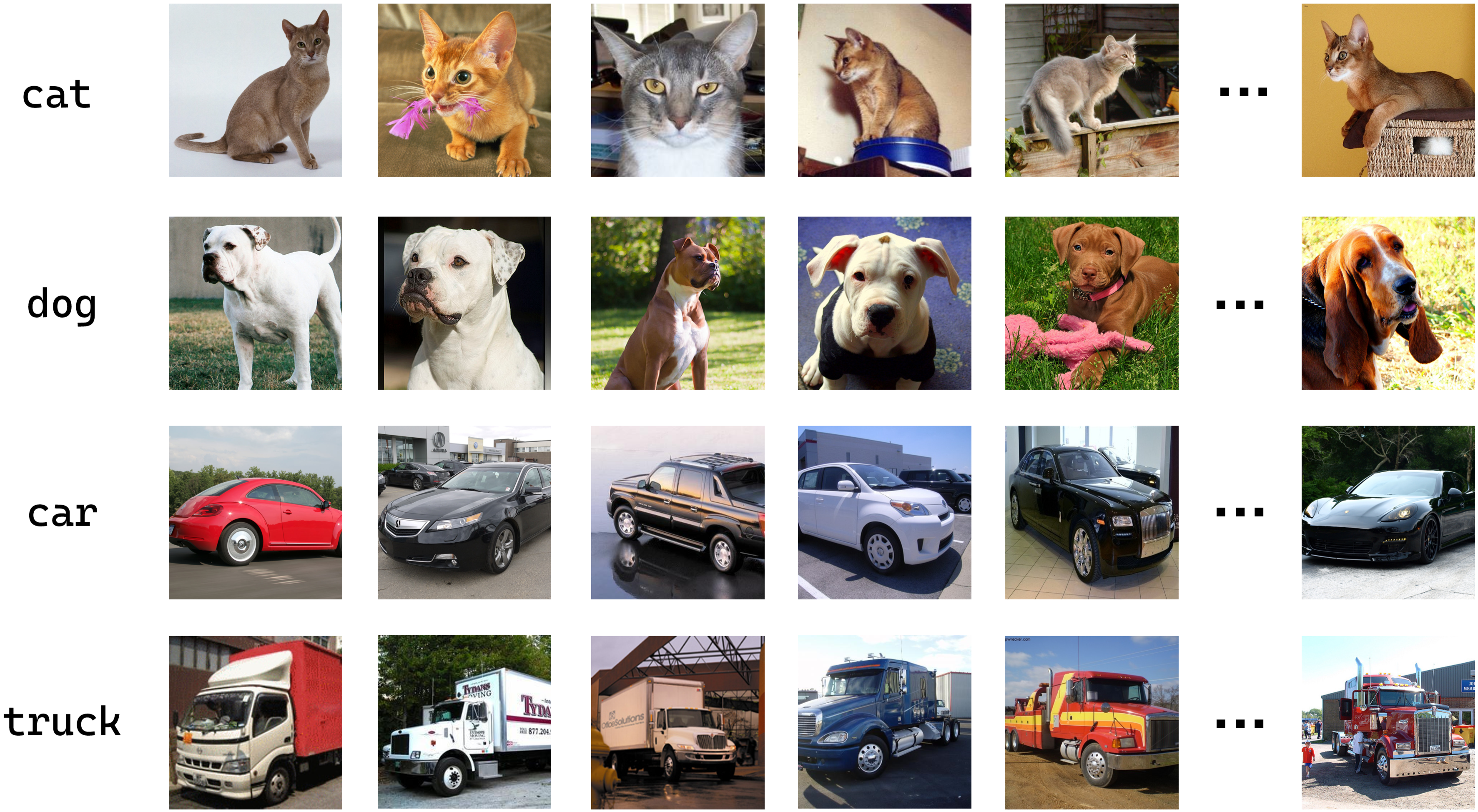}
\end{center}
\vspace{-20pt}
\caption{Top-$K$ pseudo-labeled images with the highest confidence level for 4 classes in ImageNet: \texttt{cat}, \texttt{dog}, \texttt{car}, and \texttt{truck}.}
\label{fig:topk}
\vspace{-17pt}
\end{figure}

\textbf{Pseudo-Label Generation.} Utilizing the pre-trained CLIP model, we can leverage Equation (\ref{eq-clip})  and Equation (\ref{eq:pseudo_label}) to derive pseudo-labels for unlabeled samples in the target dataset. 
To prevent the model from being overwhelmed by the abundance of samples from certain categories, we propose to select the most confident top-$K$ samples per class using Equation (\ref{eq-clip}) and Equation (\ref{eq:pseudo_label}) for subsequent optimization. Given an unlabeled training set, for each class, we choose the most confident top-$K$ samples. Specifically, we compute the similarities of all the images with the class-specific text descriptions and select the top-$K$ similar samples. In our experiments, we empirically set $K=16$. We present an illustration of our selected top-$K$ images in Fig. \ref{fig:topk}.

\textbf{Class Prototype Estimation.}
We can obtain the prototype by averaging the $L2$ normalized features of these $K$ images. For each class $c$ in a $C$-class recognition problem, we generate the visual features ${f}_k$ by CLIP's visual encoder $E_v$, then take the average of them to obtain the class prototype
\begin{equation}
    {P}_{c}=\frac{1}{K}\sum_{k=1}^K{f}_k.
\end{equation}
Here, $P_c$ represents the class prototype for the particular class $c$. The prototype set for all the classes can be denoted as $P \triangleq \{P_c\}_{c=1}^C$.  

\textbf{Prototype Adapter.}
During the inference stage, the $L2$ normalized test image feature $v_{te} \in \mathbb{R}^{1 \times C}$ generated by the visual encoder $E_v$, is used to calculate the affinities \cite{ru2022learning} between the test image and prototype, denoted as
\begin{equation}
\label{eq-H}
    H=\exp \left( -\eta \left(1-\mathrm{sim}( v_{te},P^{\top})\right) \right).
\end{equation}

Here, we use an exponential function to convert the similarities into non-negative values. $\eta$ is another hyperparameter to control its sharpness. $\mathrm{sim}\left( v_{te},P \right)$ stands for the cosine similarity between test feature $v_{te}$ and prototype $P$. We denote it as:
\begin{equation}
\label{eq-sim-vmu}
    \mathrm{sim}\left( v_{te},P \right)= v_{te}P^{\top}.
\end{equation}

 Since both $v_{te}$ and $P$ are $L2$-normalized, $v_{te}P^{\top}$ stands for their cosine similarity shown in Equation (\ref{eq-H}), the affinities between the test image and prototype can be denoted as:
\begin{equation}
\label{eq-H-new}
    H=\exp \left( -\eta (1- v_{te}P^{\top}) \right).
\end{equation}

Therefore, the prediction from the prototypes can be denoted as $H$. Besides the knowledge from the prototypes, the prior knowledge of pre-trained CLIP is calculated by $v_{te} f_t^{\top}$, where $f_t$ is the text features generated from textual encoder $E_t$. By combining the predictions of CLIP and the prototype adapter, the output logits of the test image by training-free UP-Adapter can be written as:
\begin{align}
\label{logits1}
    \mathrm{logits} &= \beta H + v_{te} f_t^{\top} \notag \\
    &= \beta \exp \left( -\eta (1- v_{te}P^{\top}) \right) + v_{te} f_t^{\top},
\end{align}
where $\beta$ denotes the residual ratio.

Even though the prediction from the prototype adapter can be achieved without training, there is still room for improving its performance. Therefore, we design a prototype model with only one linear layer for better adaptation, which can be denoted as 
\begin{equation}
    Q\left( x \right) =\exp \left( -\eta (1-Wx) \right).
\end{equation}

 Similarly, the exponential function is adopted to convert the similarities into non-negative values. As shown in Fig. \ref{fig:overview}, we initialize the network with prototype $P$, which sets the model at a high starting point. Therefore, given the image feature $v$ extracted by $E_{v}$, the logits of the UP-adapter with training can be denoted as
\begin{align}
\label{eq-LOGIT}
\mathrm{logits} &= \beta Q + v f_t^{\top} \notag \\
    &= \beta \exp \left( -\eta (1- vW^{\top}) \right) + v f_t^{\top}.
\end{align}

During training, $W$ is updated by gradient descent. We use cross-entropy loss as the loss function.

% \subsection{Inference.}
% Finally, we fuse the inference from the visual feature adapter and original CLIP to better the prediction. Therefore, we combine the logits of the visual adapter and enhanced CLIP, and the total logits of the input image $v$ are calculated as 
% \begin{align}
% \label{eq-LOGIT}
% \mathrm{Logits} &= \lambda L_v+L_C \nonumber \\ 
%                 &=\lambda \exp \left( -\eta \left( 1-Wv \right) \right) +vf_t^{\top}.
% \end{align}
% where $\lambda$ is a hyperparameter that controls how much we use to combine the prediction from the query model with enhanced CLIP.

\section{Experiments}
\subsection{Image Recognition} \label{sec:fewshot}

\textbf{Datasets.} 
Following prior fine-tuning methods for CLIP \cite{zhang2022tip}, we perform a few-shot assessment on 11 commonly used datasets for image classification. These datasets encompass various categories such as generic object classification (ImageNet \cite{recht2019imagenet}, Caltech101 \cite{fei2004learning}), fine-grained object classification (OxfordPets \cite{parkhi2012cats}, StandfordCars \cite{krause20133d}, Flowers102 \cite{nilsback2008automated}, Food-101 \cite{bossard2014food}, FGCV Aircraft \cite{maji2013fine}), texture classification (DTD \cite{cimpoi2014describing}), remote sensing recognition (EuroSAT \cite{helber2019eurosat}), scene recognition (SUN397 \cite{xiao2010sun}) and action recognition (UCF101 \cite{soomro2012ucf101}). These datasets provide a comprehensive benchmark to evaluate the few-shot
learning performance of each method.

\textbf{Baselines.}
We compare our method with four baseline methods: zero-shot CLIP \cite{radford2021learning}, CoOp \cite{zhou2022learning}, Tip-Adapter \cite{zhang2022tip} and UPL \cite{huang2022unsupervised}. Therein, CoOp \cite{zhou2022learning} and Tip-Adapter \cite{zhang2022tip} are supervised prompt tuning method and adapter-based method, respectively. UPL \cite{huang2022unsupervised} is one of the first approaches in fine-tuning vision-language models in an unsupervised setting. To ensure a fair comparison, their results are taken directly from their original paper. We also report CoOp's accuracies with the best-performance setting.

\textbf{Implementation Details.}
For pseudo-label generation, we utilize ViT-B/16 as its image encoder, while we use ResNet-50 as its image encoder in the training and inference period, we exploit Transformer as its text encoder for all the processes.   
We adopt prompt ensembling defined in \cite{radford2021learning} and follow the data pre-processing methods defined in CLIP \cite{radford2021learning} for all the datasets.  
Our model is trained for 30 epochs for ImageNet and 20 epochs for the other 10 datasets.  We conduct all the experiments on one NVIDIA RTX 3090 GPU. 

\textbf{Results.}
Table \ref{Table:SOTA} presents a comprehensive comparison of our proposed method with state-of-the-art techniques across 11 datasets. The results clearly demonstrate that our approach yields substantial performance enhancements when compared to zero-shot CLIP \cite{radford2021learning}. Moreover, our method outperforms both the 4-shot supervised CoOp \cite{zhou2022learning} and Tip-Adapter \cite{zhang2022tip} methods across all datasets, and surpasses the 8-shot versions in 10 out of 11 datasets. In comparison to another unsupervised method UPL \cite{huang2022unsupervised}, our approach consistently outperforms it in the majority of cases, resulting in an average accuracy improvement of +2.35\%, which is quite significant.

\begin{table*}[!th]
\setlength{\belowcaptionskip}{-0.01cm}
\vspace{-10pt}
\centering
\caption{Performance comparisons of our proposed methods and other state-of-the-art methods on 11 datasets. We compare our unsupervised approach with: 1) original CLIP with prompt engineering~\cite{radford2021learning}; 2) supervised methods including CoOp~\cite{zhou2022learning} and Tip-Adapter~\cite{zhang2022tip}; 3) previous unsupervised UPL~\cite{huang2022unsupervised} method.}
\resizebox{\linewidth}{!}{
\begin{tabular}{l|c|>{\columncolor{gray!20}}c|c|ccc|ccc}
\toprule
Methods  & CLIP     & \textbf{Ours}    & UPL 
        & \multicolumn{3}{c|}{CoOp}  & \multicolumn{3}{c}{Tip-Adapter} \\   \cmidrule{1-10}
        Settings & -         & \textbf{Unsupervised} & \textbf{Unsupervised} & 2-shot &  4-shot & 8-shot &2-shot & 4-shot & 8-shot \\ \midrule

ImageNet     & 60.34   &  \textbf{63.58}  &  61.09   & 57.13 & 59.72 & 61.52 & 60.96 &60.98 &  61.45                                           \\
Caltech101   & 86.09   &  \textbf{91.65}    &  91.40 & 87.76 & 89.67 & 90.14 & 89.25 &89.41 &  89.94                                                  \\
DTD          & 41.61   &  \textbf{59.06}   &  55.08 & 47.48 & 54.19 & 58.65 & 49.76 & 54.14 &  57.33                                                 \\
EuroSAT      & 38.23   &  \textbf{72.49}    &  71.04 & 59.98 & 62.17 & 68.73 & 61.10  &65.30 &   66.89                                               \\
FGVCAircraft & 16.92   &  \textbf{25.04}    &  21.75 & 20.36 & 22.10 & 24.99 & 21.25 &21.54 &  24.48                                                 \\
Food101      & 77.33   &   \textbf{78.06}   &  77.93 & 72.92 & 73.74 & 76.28 & 77.58 &77.60 &  77.79                                                 \\
Flowers102   & 66.06   &   86.19   &  76.65 & 76.58 & 84.59 & \textbf{88.27} & 76.82 &81.53 &  85.95                                                 \\
OxfordPets   & 85.83   &   \textbf{89.66}   &  89.51 & 84.53 & 87.11 & 87.71 & 87.38 &87.67 &  87.87                                                 \\
SUN397       & 60.18   &   \textbf{68.52}   &  66.42 & 61.35 & 65.08 & 67.47 & 62.82 &64.32 &  65.57                                                \\
StandfordCars& 55.64   &   68.31  &  \textbf{70.97} & 59.49 & 61.92 & 65.25 & 59.86 &62.03 &  63.35                                                \\
UCF101       & 62.70   &   \textbf{75.35}   &  70.18 & 65.06 & 68.26 & 71.67 & 66.59 &67.51 &  69.10                                                 \\
\midrule
\textbf{Average} & 59.18 & \textbf{70.72} &68.37& 62.97 & 66.23   &  69.15 &  64.85 & 66.55 & 68.16           \\ \bottomrule
\end{tabular}
}
\vspace{-27pt}
\label{Table:SOTA}
\end{table*}

\subsection{Domain Generalization}

\textbf{Experiment Setup.} It is critical for the machine learning model to have strong robustness to distribution shifts \cite{tang2023cross,tang2023neuro}.
To demonstrate the generalization capabilities of our proposed method, We evaluate the performance on the domain generalization task. We follow prior methods to train our model on 16-shot ImageNet \cite{deng2009imagenet} and test on four ImageNet variant datasets (ImageNet-V2 \cite{recht2019imagenet}, ImageNet-Sketch \cite{wang2019learning}), ImageNet-A \cite{hendrycks2021natural}, and ImageNet-R \cite{hendrycks2021many}.

\textbf{Baselines.}
We include six previous methods for comparison: zero-shot CLIP \cite{radford2021learning}, linear probe CLIP \cite{radford2021learning}, CoOp \cite{zhou2022learning}, CoCoOp \cite{zhou2022conditional}, and TPT \cite{shu2022tpt}. All of these state-of-the-art methods are supervised methods. For a fair comparison,
we directly include the results of other baselines from their original paper.

\textbf{Results.}
Table \ref{table:generalization} presents the performance comparison results on domain generalization tasks (from ImageNet to ImageNet-V2/-Sketch/-A/-R). Our proposed unsupervised UP-Adapter outperforms the supervised baselines in accuracy, which suggests remarkable robustness to distribution shifts of our method. 

\begin{table}[htbp]
% \vspace{-15pt}
\setlength{\belowcaptionskip}{-0.2cm}
\small
\begin{center}
\caption{Comparison with other methods on domain generalization task. The best results are in \textbf{bold}.}
\label{table:generalization}
\resizebox{\linewidth}{!}{
\begin{tabular}{lccccccc}
\toprule
\multirow{2}*{Method}  & \multirow{2}*{Unsupervised?} &Source & \multicolumn{5}{c}{Target} \\ \cmidrule(lr){3-3} \cmidrule(lr){4-8} & & ImageNet & -V2 & -Sketch & -A & -R  & OOD Average \\
% \multirow{2}*{Method} & \multirow{2}*{Backbone} & Source &\multicolumn{4}{c}{Target} \\  \cline{4-6} & ImageNet & ImageNet-V2 &  ImageNet-Sketch & Average & OOD Average \\
\midrule
% \textbf{ResNet-50} & \\
Zero-Shot CLIP~\cite{radford2021learning} &- &  60.33 &  53.27  & 35.44  & 21.65 &  56.00  & 41.59\\
Linear Probe CLIP~\cite{radford2021learning} &-  & 56.13  & 45.61  & 19.13  & 12.74  & 34.86 &  28.09\\
CoOp~\cite{zhou2022learning} &\xmark   & 63.33  & 55.40  & 34.67  & 23.06  & 56.60 &  42.43\\
CoCoOp~\cite{zhou2022conditional} &\xmark   & 62.81  & \textbf{55.72}  & 34.48  & 23.32  & 57.74  & 42.82\\
TPT~\cite{shu2022tpt} &\xmark &   60.74  & 54.70  & 35.09  & 26.67  & 59.11  & 43.89\\
\rowcolor{gray!20}
\textbf{Ours} &\cmark  & \textbf{63.58}  & 54.90  & \textbf{35.56}  & \textbf{27.65}  & \textbf{59.36}  & \textbf{44.37}\\
\bottomrule
\end{tabular}
}
\vspace{-18pt}
\end{center}
\end{table}

\subsection{Ablation Study}
\begin{wraptable}[9]{r}{0.58\textwidth}
\vspace{-40pt}
\small
\begin{center}
\caption{Effects of different algorithm components. We report the average accuracies on 11 image recognition datasets.}
\vspace{-5pt}
\label{table:abl}
\resizebox{\linewidth}{!}{
\begin{tabular}{lc}
\toprule
Methods & Accuracy \\
% \multirow{2}*{Method} & \multirow{2}*{Backbone} & Source &\multicolumn{4}{c}{Target} \\  \cline{4-6} & ImageNet & ImageNet-V2 &  ImageNet-Sketch & Average & OOD Average \\
\midrule
% \textbf{ResNet-50} & \\
Zero-shot CLIP~\cite{radford2021learning} &59.18\\
Prototype adapter only &66.70\\
CLIP + class prototypes &67.31\\
CLIP + prototype adapter (w/o initialization) &65.18\\
\rowcolor{gray!20}
CLIP + prototype adapter (w/ initialization) &70.72\\
\bottomrule
\end{tabular}
}
\vspace{-20pt}
\end{center}
\end{wraptable}
\textbf{Effects of Different Algorithm Components.} To better demonstrate the effectiveness of our UP-Adapter, we have conducted an ablation study on the ImageNet \cite{deng2009imagenet} dataset to study the effects of different algorithm components. The results are shown in Table \ref{table:abl}. As we can see from the table, directly using the class prototypes is not fully effective. The initialization of the prototype is also crucial, which will enhance the performance by 5.54\%.

\textbf{Number of Selected Images for Each Class $K$.}
In our approach, we carefully choose the top-$K$ most confident samples for the prototype estimation process. In our experiments, we empirically set $K=16$. In Table \ref{table:ablation}, we show an ablation study on the number of selected images for each class $K$. We report the average accuracies over 11 datasets on image recognition tasks using $K= 4, 8, 16, 32$ to investigate the impact of $K$.  This ablation study suggests that our setting of $K=16$ yields the optimal performance.

\textbf{Visual Backbones for Pseduo-Labeling.}
Table \ref{table:ablation} provides a summary of average accuracies across 11 datasets using different visual backbones, including ResNets and ViTs. Notably, employing more advanced backbones results in more accurately generated pseudo-labels, which in turn leads to improved overall performance. When using the ViT-B/16 backbone, we achieve the best average accuracy of 70.72\%.

\begin{table}[htbp]
\vspace{-10pt}
\setlength{\tabcolsep}{6pt}
\caption{Parametric analysis on values of $K$ and visual backbones.}
\vspace{-5pt}
\centering
\resizebox{\textwidth}{!}{
	\begin{tabular}{cccc|cccc}
 
	\toprule
		\multicolumn{4}{c|}{Number of Selected Images $K$}  & \multicolumn{4}{c}{Visual Backbones for Pseduo-Labeling} \\
  \cmidrule(lr){1-4}\cmidrule(lr){5-8}
        $K=4$ &$K=8$ &$K=16$ &$K=32$ & ResNet-50  & ResNet-101  &ViT-B/32 &ViT-B/16 \\ 
		 67.03  & 68.97  &\textbf{70.72} &70.21 & 65.44  &67.26 &68.81 &\textbf{70.72}\\ 
	\bottomrule
	\end{tabular}
 }
\vspace{-15pt}
\label{table:ablation}
\end{table}

\section{Conclusion}
In this work, we propose an unsupervised fine-tuning approach for vision-language models to tackle the challenge of requiring a substantial amount of annotated data. We utilize CLIP's capabilities to generate pseudo-labels for unlabeled samples, and we carefully choose the most confident samples for each class. Using these selected samples, we create prototypes specific to each class, which serve as initializations for an adapter module. This initialized adapter module can be further fine-tuned to handle recognition tasks effectively.
Extensive experimentation in the domains of few-shot image recognition and domain generalization reveals that our proposed unsupervised method outperforms the state-of-the-art unsupervised UPL method by substantial margins.

%
% ---- Bibliography ----
%
% BibTeX users should specify bibliography style 'splncs04'.
% References will then be sorted and formatted in the correct style.
%
\bibliographystyle{splncs04}
\bibliography{prcv2023}
\end{document}